\pdfoutput=1

\documentclass[11pt]{article}

\PassOptionsToPackage{table,xcdraw}{xcolor}
\usepackage[final]{acl}
\usepackage{authblk}
\usepackage{times}
\usepackage{latexsym}
\usepackage{multirow} 
\usepackage{booktabs}
\usepackage{pifont}

\usepackage[T1]{fontenc}

\usepackage[utf8]{inputenc}

\usepackage{microtype}

\usepackage{inconsolata}

\usepackage[table]{xcolor}
\usepackage{tcolorbox}
\usepackage{pifont}

\usepackage{algorithm}
\usepackage{algorithmic}
\usepackage{amssymb}
\newcommand{\cmark}{\ding{51}}%
\newcommand{\blueright}{\textcolor{blue}{\cmark}}

\newcommand{\xmark}{\ding{55}}%
\newcommand{\redwrong}{\textcolor{red}{\xmark}}
\usepackage{graphicx}
\usepackage{newfloat}
\usepackage{float}
\usepackage{listings}
\DeclareCaptionStyle{ruled}{labelfont=normalfont,labelsep=colon,strut=off} 
\usepackage{amsmath}

\title{\includegraphics[scale=0.037]{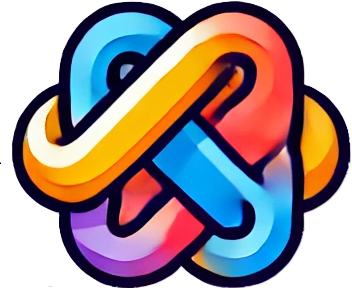} Reefknot: A Comprehensive Benchmark for Relation\\ Hallucination Evaluation, Analysis and Mitigation in Multimodal Large Language Models}

\author{
Kening Zheng$^{1 *}$ , Junkai Chen$^{1 }\thanks{Equal Contribution.}$, Yibo Yan$^{1,3}$, Xin Zou$^{1}$, Xuming Hu$^{1,2,3}$\thanks{Corresponding author.} \\
$^{1}$Hong Kong University of Science and Technology (Guangzhou) \\
$^{2}$Guangxi Zhuang Autonomous Region Big Data Research Institute \\
$^{3}$Hong Kong University of Science and Technology \\
\texttt{\{neok2zkn, junkai.chen.0917\}@gmail.com} \\
\texttt{ xuminghu@hkust-gz.com}
}

\begin{document}
\maketitle

\begin{abstract}
Hallucination issues continue to affect multimodal large language models (MLLMs), with existing research mainly addressing object-level or attribute-level hallucinations, neglecting the more complex relation hallucinations that require advanced reasoning. Current benchmarks for relation hallucinations lack detailed evaluation and effective mitigation, and their datasets often suffer from biases due to systematic annotation processes.
To address these challenges, we introduce Reefknot, a comprehensive benchmark targeting relation hallucinations, comprising over 20,000 real-world samples. We provide a systematic definition of relation hallucinations, integrating perceptive and cognitive perspectives, and construct a relation-based corpus using the Visual Genome scene graph dataset. Our comparative evaluation reveals significant limitations in current MLLMs' ability to handle relation hallucinations.
Additionally, we propose a novel confidence-based mitigation strategy, which reduces the hallucination rate by an average of 9.75\% across three datasets, including Reefknot. 
Our work offers valuable insights for achieving trustworthy multimodal intelligence.
The dataset and code are released at \url{https://github.com/JackChen-seu/Reefknot}.
\end{abstract}
\section{Introduction}
\begin{table*}[t]
\centering
\small
\setlength{\tabcolsep}{4.0pt}
\begin{tabular}{lccccccc}
\toprule
\multirow{2}{*}{\textbf{Benchmarks}} & \multicolumn{2}{c}{\textbf{Dataset}} &\multicolumn{3}{c}{\textbf{Evaluation}} & \multicolumn{2}{c}{\textbf{Analysis}}   \\ 
\cmidrule(lr){2-3} \cmidrule(lr){4-6} \cmidrule(lr){7-8}
 & \textbf{Source} &\textbf{Construction}&  \textbf{Y/N} & \textbf{MCQ} & \textbf{VQA} & \textbf{Metric}& \textbf{Improv. Focus} \\
 \midrule
POPE&COCO&Post-processed &\blueright &\redwrong &\redwrong& Acc. &Co-occur.\\
HaELM&MS-COCO&Post-processed &\blueright&\redwrong&\redwrong&Acc. &Attention\\
MME&Self-Sourced&Manual&\blueright&\redwrong&\redwrong&Acc.&-\\
AMBER&MS-COCO&Post-processed &\blueright&\redwrong&\redwrong&Acc.&-\\
MHaluBench&Self-Sourced&Post-processed &\blueright&\redwrong&\redwrong&Prec.&-\\
R-Bench&COCO&Post-processed &\blueright&\redwrong&\redwrong&Acc.&Co-occur.\\
MMRel&Visual-Genome&Multi-source&\blueright&\redwrong&\redwrong&Acc.&-\\
VALOR-EVAL&GQA&Post-processed &\redwrong&\redwrong&\redwrong&LLM-based&Co-occur.\\
FAITHSCORE&MS-COCO \& LLaVA-1k&Post-processed &\redwrong &\redwrong &\redwrong& Faithscore &Atomic Facts.\\
FIHA&MS-COCO Foggy \& VG&Ruled-based &\blueright&\redwrong&\blueright&Acc. &Annotation Free\\
TIFA&Synthetic&Caption-based&\blueright&\blueright&\redwrong &Acc.&T2I Evaluation\\
\midrule
\cellcolor{gray!8}{{Reefknot (Ours)}} & \cellcolor{gray!8}Visual-Genome & \cellcolor{gray!8}Original&\cellcolor{gray!8}\blueright&\cellcolor{gray!8}\blueright&\cellcolor{gray!8}\blueright&\cellcolor{gray!8}$R_{socre}$&\cellcolor{gray!8}Confidence\\ \bottomrule
\end{tabular}
\caption{Comparsions of our proposed Reefknot benchmark with relevant benchmarks. POPE~\cite{POPE}, HAELM~\cite{HaELM}, MME~\cite{MME}, AMBER~\cite{AMBER}, MHaluBench~\cite{MHaluBench}, R-bench~\cite{RBench}, MMRel~\cite{MMREL}, VALOR-EVAL~\cite{VALOR-EVAL}, FAITHSCORE~\cite{faithscore}, FIHA~\cite{FIHA}, TIFA~\cite{TIFA}}
\label{tab:bench comparison}
\end{table*}
In recent years, large language models (LLMs) have revolutionized the AI field by expanding their training data to trillions of tokens and increasing their parameter counts to hundreds of billions~\cite{brown2020language, achiam2023gpt, touvron2023llama}. 
This has unlocked powerful emergent abilities, and seen widespread applications in diverse domains~\cite{achiam2023gpt,yan2024urbanclip,wang2023surveyfactualitylargelanguage}. 
Recently, the community managed to combine visual backbones with powerful LLMs, resulting in multimodal large language models (MLLMs)~\cite{liu2023llava}. 
While this has led to advancements in multimodal scenarios, it also presents challenges for MLLMs, notably their tendency to generate hallucinations \cite{zou2024look}.
In LLMs, hallucinations occur when the model generates inaccurate or misleading factual information that cannot be supported by existing knowledge~\cite{siren}. 
However, such issues become more complex in MLLMs, as hallucinations can manifest as responses containing references or descriptions of the input image that are incorrect \cite{bai2024hallucination,huo2024mmneuron}.
Therefore, it is crucial to evaluate and mitigate these hallucinations to improve the trustworthy of MLLMs in real-world scenarios.

\begin{figure}[!t]
    \centering
    \includegraphics[scale=0.32]{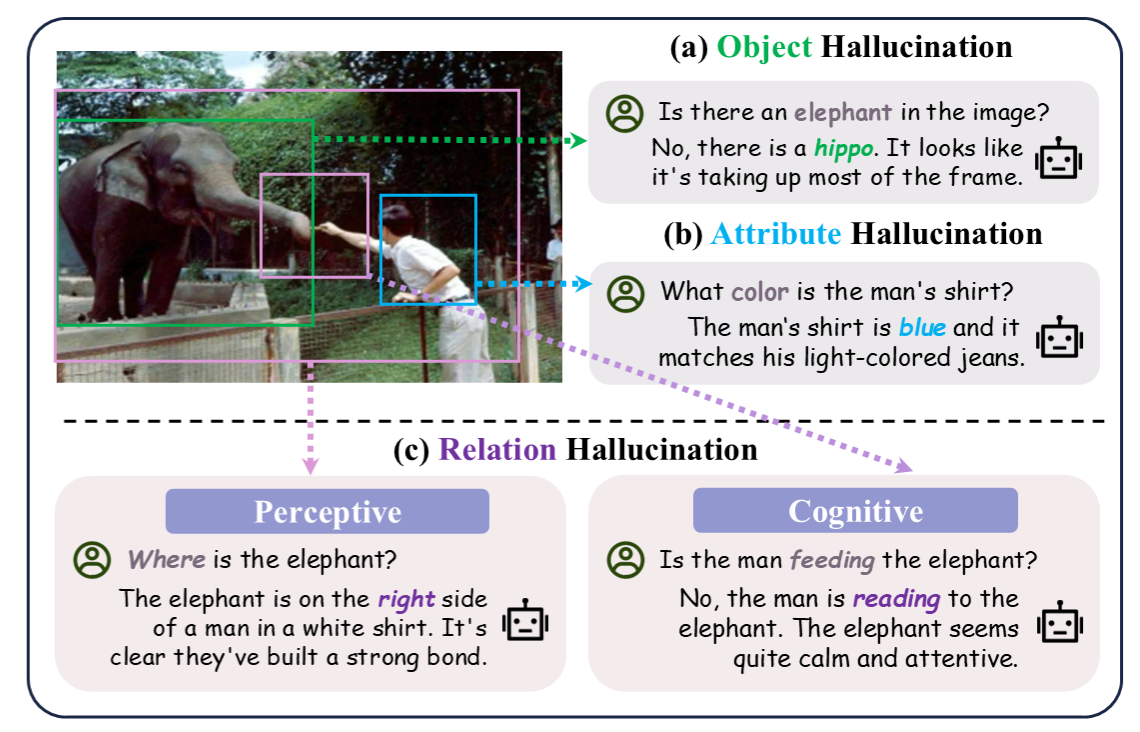}
    \caption{Comparison between the focus of Reefknot — relation hallucination with two categories (i.e., perceptive \& cognitive) vs. object \& attribute hallucinations.}
    \label{fig:teaser}
\end{figure}
Hallucination in MLLMs likely originates from knowledge biases between pre-training and fine-tuning, including statistical biases in the training data, over-reliance on parametric knowledge, and skewed representation learning, as suggested by previous research \cite{bai2024hallucination,zhu2024fastmem}.
Specifically, the hallucination in MLLMs can be divided into three categories: \textbf{\textit{object}}, \textbf{\textit{attribute}} and \textbf{\textit{relation}} hallucinations \cite{bai2024hallucination}. 

As depicted in Figure \ref{fig:teaser} (a), object-level hallucination focuses on the model's discrimination of the existence of basic objects~\cite{POPE}; while as shown in Figure \ref{fig:teaser} (b), attribute-level hallucination often focuses on whether the model can distinguish some properties of the object itself like color, number, shape and so on \cite{MME}. Their commonality is that they only focus on the single object present in the image. There has been much work exploring the alleviation of these two types of hallucinations. For example, Woodpecker~\cite{woodpecker} used a post-processing method to correct hallucination after the generation process; VCD~\cite{vcd} proposed a contrast decoding strategy to mitigate object-level hallucination by adding noise to images.

\begin{figure}[!t]
    \centering
    \includegraphics[scale=0.38]{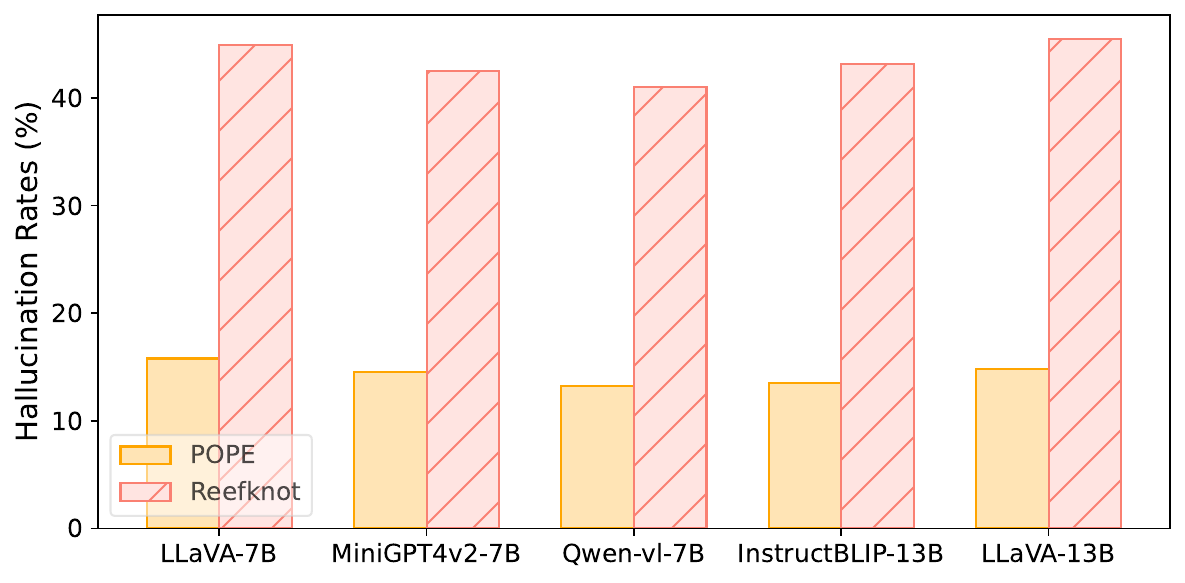}
    \caption{The hallucination rates on POPE, an object hallucination benchmark, and our Reefknot with a focus on relation hallucination (w/ same configuration).}
    \label{fig:domain}
\end{figure}

Despite these efforts, \textit{the community barely considers relation hallucination, which demands more complex reasoning capabilities from MLLMs}. 
Figure \ref{fig:teaser} (c) reveals that relation hallucination is related to at least two objects simultaneously in the image, through either perceptive or cognitive perspective. 
Furthermore, as outlined in Table \ref{tab:bench comparison}, recent highly relevant benchmarks on hallucinations \textit{lack thorough evaluation and effective mitigation strategies}.

Specifically, these works either adopted either simple Yes/No (Y/N) task to assess the models' accuracy/precision, or utilized LLM to score the performance, but none of them were able to give a comprehensive evaluation from both discriminative (Y/N MCQ) and generative (VQA) perspectives.

Moreover, previous benchmarks seldom proposed mitigation methods, with only a few focusing on co-occurrence or attention mechanisms to address these issues. In contrast, our paper analyzes token-level confidence at each layer to detect and promptly mitigate hallucinations.

To handle the aforementioned research gaps, we present a specific definition for relation hallucination and propose the first comprehensive benchmark \textbf{Reefknot} to evaluate the performance on \textit{relation hallucination}.
\textbf{\textit{Relation hallucination}} \emph{refers to the phenomenon in which MLLMs misinterpret the logical relationships between two or more objects in  the corresponding image}.
Unlike many benchmarks of MLLM that were constructed by automatic labeling technique, we construct the relation-based dataset based on semantic triplets retrieved from the scene graph dataset. 
Table \ref{tab:bench comparison} demonstrates that our triplets are from real-life scenarios, without any post-processing (e.g., segmentation and bounding box techniques), manual annotation, and synthetic method (e.g., diffusion-based generation). 
We categorize relation hallucinations into two types: perceptive, involving concrete relational terms like \textit{"on", "in", "behind"} and cognitive, which encompasses more abstract terms such as \textit{"blowing"} and  \textit{"watching"}. Second, we evaluate the mainstream MLLMs on Reefknot via three diverse tasks across two types of relation hallucinations. Figure \ref{fig:domain} also illustrates that relation hallucination can be more severe than object hallucination in current MLLMs, highlighting the importance of our evaluation. Furthermore, we propose a simple relation hallucination mitigation method named Detect-then-Calibrate.
This originates from the experimental observation that when relation hallucinations occur, the response probability drops significantly, hovering just above 50\% in extreme cases compared to the usual nearly 90\%. 
Our method achieves an average improvement of 9.75\% across three relation hallucination benchmarks.
In summary, Our main contributions are as follows:

\begin{enumerate}
    \item We have constructed Reefknot, a benchmark comprising two types of relationships and three evaluation tasks to assess relation hallucinations comprehensively. 
    \item We have conducted a thorough evaluation of relation hallucinations across mainstream MLLMs, uncovering that these models are disproportionately susceptible to perceptual hallucinations in comparison to cognitive ones.
    \item We investigated the mechanism of relation hallucination generation from the perspective of response confidence and identified a correlation between relation hallucination and high uncertainty from token levels. 
    \item We have proposed a novel Detect-then-Calibrate method to detect and mitigate hallucination. By analyzing token confidence scores, we established a threshold to identify hallucinations. Further, we applied a calibration strategy to mitigate hallucination at intermediate confidence levels. Extensive experiments on three relation hallucination datasets demonstrate the effectiveness of our approach.
\end{enumerate}

\section{Relation Hallucination Benchmark}

\begin{figure*}[t]
\centering
\includegraphics[width=1.01\textwidth]{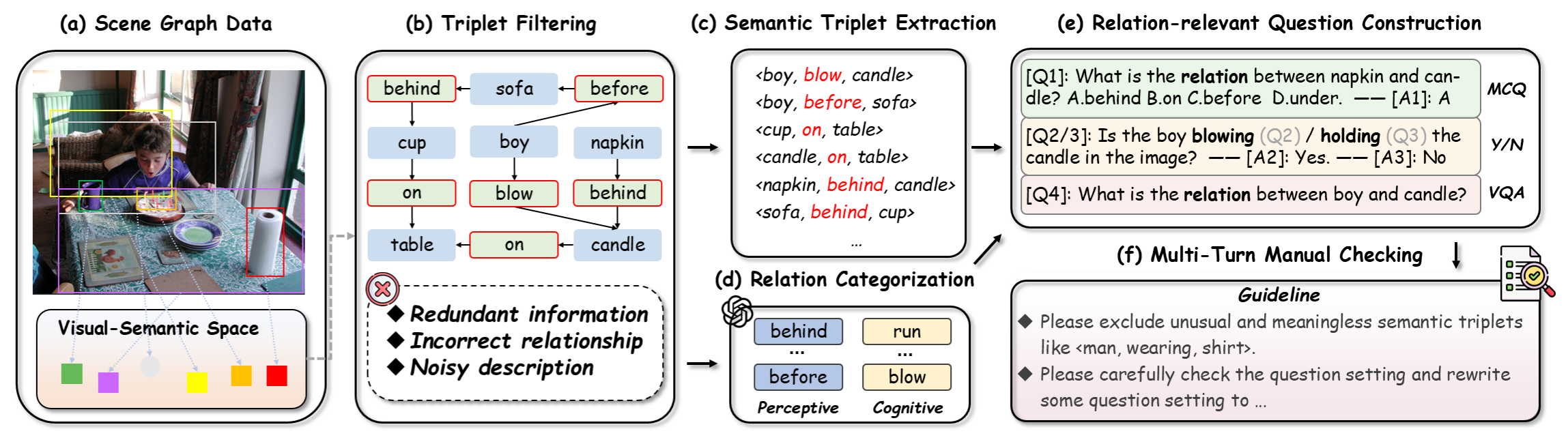}
\caption{The data construction pipeline of our proposed Reefknot benchmark. }
\label{fig:flow}
\end{figure*}

In this section, we describe the dataset construction pipeline of Reefknot benchmark in Figure \ref{fig:flow}. Unlike object and attribute hallucinations that only involve one entity, relations involve two entities making it more difficult to handle. We first identify relation triplets from Visual Genome (VG) dataset \cite{vg} (Phase a), and conduct triplet filtering (Phase b). Subsequently, we extract the semantic triplets (Phase c) and categorize their relations (Phase d). Then, a relation-based question set can be constructed into three types (Phase e). Finally, the quality of dataset is ensured by three rounds of expert-based validation (Phase f).

\paragraph{Triplet Identification, Filtering and Extraction} The dataset comprises of 11,084 images taken from VG dataset, a finely annotated scene graph dataset utilized by the research community \cite{tang2020unbiased,vRr-VG}. As indicated in Figure \ref{fig:flow} (a), visual objects and their relations from VG dataset can be easily identified. Besides, we filter the triplets with redundant information, incorrect relationships, or noisy descriptions, as depicted in Figure \ref{fig:flow} (b). Subsequently, we can extract semantic triplets by identifying subject-object pairs and the relationships between them, forming \textit{(subject, relation, object)} triplets in Figure \ref{fig:flow} (c).

\paragraph{Relation Categorization} As depicted in Figure \ref{fig:flow} (d), we categorize relationships into two categories based on deep semantic meanings: perceptive and cognitive. Perceptive relationships involve locational prepositions, such as \textless boy, behind, sofa \textgreater and \textless cup, on, table\textgreater; whereas cognitive relationships are expressed through action phrases indicating states, such as \textless boy, eating, food\textgreater and \textless girl, sleeping in, bed\textgreater. ChatGPT is employed to assist in this classification. Table \ref{tab:dataset_info} also indicates the sample numbers in different tasks and hallucination categories. The prompt we used for relation categorization is listed in Appendix \ref{section prompt}.

\begin{table}
\centering
\small
\setlength{\tabcolsep}{9pt}
\begin{tabular}{lcccc}
\toprule
\textbf{Category} & \textbf{Y/N} & \textbf{MCQ} & \textbf{VQA} & \textbf{Total} \\
\midrule
\#Perception & 5,440 & 4,800 & 2,150 & 13,260 \\
\#Cognition & 4,300 & 2,150 & 2,720 & 8,600 \\
\#Total & 9,740 & 6,950 & 4,870 & 21,880 \\
\midrule
\midrule
\multicolumn{4}{l}{Ratio of positive and negative samples} & 1:1 \\
\multicolumn{4}{l}{Number of perceptive relationship types} & 56 \\
\multicolumn{4}{l}{Number of cognitive relationship types} & 152 \\

\multicolumn{4}{l}{Number of images} & 11084 \\
\bottomrule
\end{tabular}
\caption{Detailed statistical overview of the Reefknot benchmark dataset, including sample distribution across categories and task types.}
\label{tab:dataset_info}
\vspace{-2mm}
\end{table}
\begin{table*}[htbp]
\small
\setlength{\tabcolsep}{7pt}
    \centering
    \begin{tabular}{cccccccccc}
        \toprule
        \multicolumn{1}{c}{\multirow{2}{*}{\includegraphics[scale=0.35]{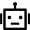} MLLMs\quad}} & \multirow{2}{*}{Size}&\multicolumn{3}{c}{Perception$\downarrow$} & \multicolumn{3}{c}{Cognition$\downarrow$} & Total$\uparrow$ \\
          \cmidrule(lr){3-5} \cmidrule(lr){6-8} \cmidrule(lr){9-9} 
          & &Y/N &MCQ & VQA &  Y/N &MCQ & VQA&$R_{score}$\\
         \midrule
         \multicolumn{1}{l}{Phi-3-vision-128k-instruct~\cite{phi3}}&4.2B&39.88&57.07&50.98&33.97&21.35&49.45& 60.30 \\
         \multicolumn{1}{l}{Yi-VL~\cite{Yi34b}}&6B&33.56&47.53&71.02&33.16&16.33&74.96&55.81  \\
         \multicolumn{1}{l}{LLaVA~\cite{liu2023improved}}&7B&37.67&68.05&52.93&33.99&51.04&54.56&51.41  \\
        
        \multicolumn{1}{l}{MiniGPT4-v2~\cite{minigptv2}}&7B& 46.7 & 78.00 & 61.30 & 43.73 & 68.50 & 65.88 & 39.88  \\
  
       \multicolumn{1}{l}{MiniCPM~\cite{MiniCPM}}&7B& \textbf{31.93} & 48.65 & \textbf{47.63} & 27.65 & 16.71 & \textbf{45.96} & \textbf{65.73}  \\
        \multicolumn{1}{l}{Qwen-VL~\cite{qwen}}&7B& 42.21 & 56.7 & 72.47 & 33.53 & 21.88 & 73.01 & 52.55  \\
       \multicolumn{1}{l}{Deepseek-VL~\cite{deepseek}}&7B& 37.58 & 56.33 &67.07  & 32.22 & 23.60 & 59.34 & 56.39  \\
       \multicolumn{1}{l}{GLM4V~\cite{ChatGLM}}&9B& 34.09 & 50.47 & 58.09 & \textbf{27.08} & 16.87 & 56.47 & 62.03  \\
        \multicolumn{1}{l}{LLaVA~\cite{liu2023improved}}&13B& 40.7 & \underline{59.35} & 48.93 & 34.19 & \underline{29.19} & 54.45 & 57.47  \\
        \multicolumn{1}{l}{CogVLM~\cite{Cogvlm}}&19B& 37.23 & 47.95 & 70.14 & 29.89 & 18.54 & 66.18 & 57.1  \\
       \multicolumn{1}{l}{Yi-VL~\cite{Yi34b}}&34B& 32.79 & \textbf{44.19} & 57.67  & 33.75 & \textbf{14.85} & 52.72 & 62.61  \\ \midrule
       \multicolumn{1}{l}{GPT-4o~\cite{gpt4o}}&-& 32.56 & 40.93 &42.70  & 26.27 & 11.53 & 48.78 & 68.32  \\
       \bottomrule
    \end{tabular}
    \caption{\textbf{Evaluation of hallucination rates} on the different MLLMs. Additionally, we use \textbf{bold} to highlight the best performance of open-sourced MLLMs, and \underline{underline} to emphasize the distinction between perception and cognition of LLaVA-13B. Note all experiments are done with the temperature to 0 to keep the reproducibility.}
    \label{main results}
\end{table*}

\paragraph{Relation-Relevant Question Construction} As shown in Figure \ref{fig:flow} (e), we construct three types of relation-relevant question sets to evaluate the state-of-the-art MLLMs' abilities in relation-level perception and reasoning.
\begin{itemize}
    \item For \texttt{Yes/No} (Y/N) questions, we employ an adversarial approach by introducing a negative sample within the same triplet, alongside the positive sample, to test the model's ability to correctly answer "No".
    \item \texttt{Multiple Choice Questions} (MCQ) are designed with one correct answer and three random options to evaluate the model's resistance to relation hallucinations within a controlled and limited vocabulary.
    \item \texttt{Visual Question Answering} (VQA) is an open-ended task that allows us to comprehensively assess a model's instruction-following capabilities and relation perception within an open-domain environment.
\end{itemize}

\paragraph{Multi-Turn Manual Checking} Finally, we perform multi-turn manual verification to ensure the quality of the question sets (see Figure \ref{fig:flow} (f)). Each question undergoes at least three rounds of review by four domain experts. We revise any inappropriate expressions and exclude meaningless questions, such as \textit{``Is the window on the wall?"}, which lack informative value and can be answered without visual input. 
After rigorous screening process, our dataset comprises 21,880 questions across 11,084 images as shwon in Table \ref{tab:dataset_info}.

\begin{figure}[ht]
    \centering
    \includegraphics[scale=0.4]{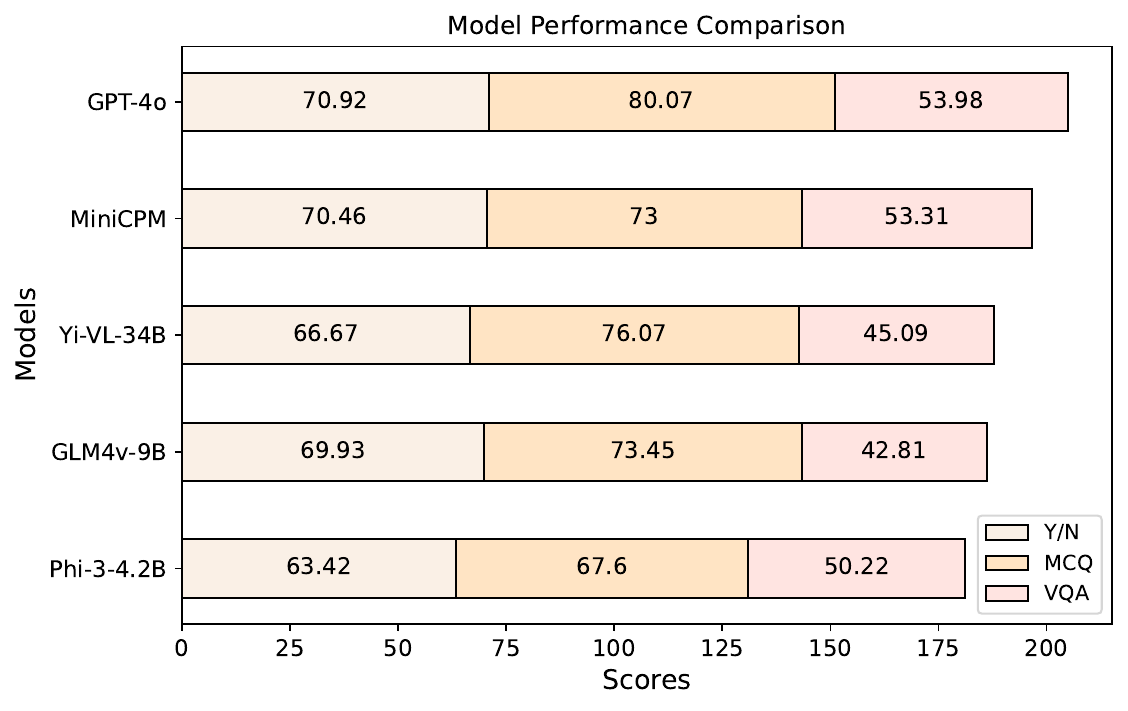}
    \caption{MLLMs with \textbf{top five best performance evaluated on Reefknot benchmark}. We report the sum of the respective metric across three tasks for reference.}%
    \label{fig:top5}
\end{figure}

\section{Hallucination Evaluation}
\subsection{Task and Evaluation Metrics}
As illustrated in Table \ref{main results}, we conducted evaluations on mainstream MLLMs to evaluate relation-level hallucinations.
All experiments of open-sourced MLLMs were conducted using 8 NVIDIA A100 GPUs. 
Each experiment ran three times, and we reported the average of these results.
In our evaluation, we reported our results from two distinct categories: perception and cognition.

For discriminative Y/N and MCQ tasks, we reported the hallucination rate $Halr$ as a metric.
Generative questions have always posed a challenge in evaluation of MLLMs.
In our assessment of the generative VQA task, we employed the DeBERTa model using a bidirectional entailment approach for label match~\cite{SU}, and we denoted $Halr$ as metric for simplicity as well.
For more details, we have listed in Appendix \ref{vqaCriterion}.
In general, we use the following metric $R_{\text{score}}$ to comprehensively evaluate the overall performance across the three tasks, which is expressed in Formula \ref{formula1}.

\begin{equation}
\label{formula1}
    R_{\text{score}} = \text{Avg} \left[ \sum_{i=1}^{3} (1 -Halr_{i}) \right].
\end{equation}

\subsection{Main results} 

\paragraph{Overall Performance} Significant performance differences among the various models are evident.
Table \ref{main results} shows significant differences in the performance of various models across different question types.
For instance, Qwen-vl-chat excels in Y/N and MCQ settings but encounters serious hallucination issues in VQA tasks.
A detailed review of the model's responses reveals that while Qwen can follow instructions accurately, many responses contain expressions that are completely unrelated to the labels. 
We presume that such a phenomenon is owing to over-fitting training during the instruction-tuning process.
Among open-sourced models, MiniCPM~\cite{MiniCPM} stands out, likely due to its adoption of the fine-grained alignment technique such as RLHF-V~\cite{rlhf-v} to alleviate hallucinations.

\begin{figure}[ht]
    \centering
    \includegraphics[scale=0.5]{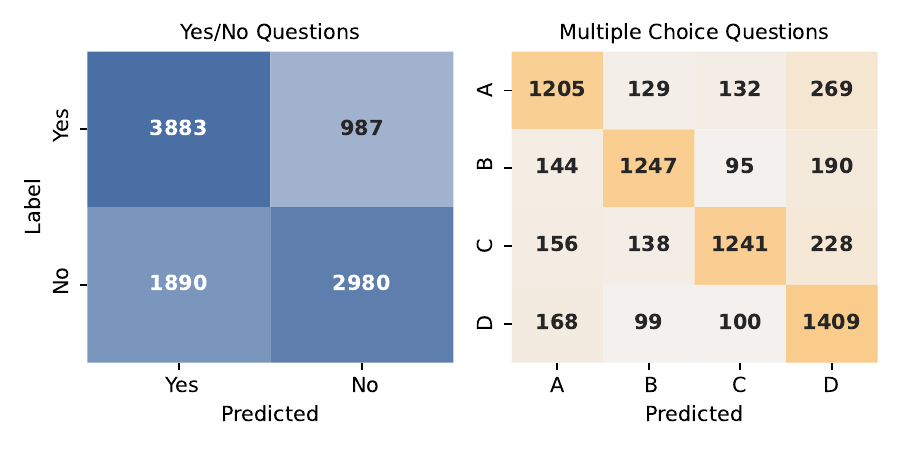}
    \caption{\textbf{Confusion matrices} of MiniCPM-7B's performance on Reefknot benchmark (Left: Y/N setting; Right: MCQ setting). }%
    \label{confusion}
\end{figure}
\paragraph{Error Analysis} 
Figure~\ref{fig:top5} presents a comparative analysis of the top five best models' performances on the Reefknot benchmark. The comparison reveals a clear hierarchy in performance, with GPT-4o outperforming the rest across all tasks. MiniCPM, Yi-VL-34B, and GLM4v-9B show a competitive edge, closely following GPT-4o, while Phi-3 lags behind relatively. In terms of task performance, Figure~\ref{fig:top5} indicates that while some models excel in one task, they may not perform as well in others. For example, Yi-VL-34B achieves the second-highest score in MCQ but underperforms in VQA. This underscores the significance of task-specific model tuning and the necessity of a balanced approach to enhance overall performance across all relation hallucination tasks.
\paragraph{Perceptive vs. Cognitive Hallucination} We find the occurrence of cognitive hallucinations is generally lower than that of perceptive hallucinations, which may diverge from intuition.
Across all models and settings, the incidence of perceptive hallucinations is consistently 10\% higher than that of cognitive hallucinations.
In the most extreme case, such as LLaVA-13B model in the MCQ setting, the rate of perceptive hallucinations is 30.16\% higher than that of cognitive hallucinations.
This phenomenon may be caused by the fact that models often utilize large-scale image-caption datasets during the pre-training and fine-tuning processes.
These datasets typically contain detailed visual descriptions, enabling models to perform better in cognitive relationships such as \textit{running, eating,} etc.
Conversely, these models may struggle when dealing with some perceptive relationships based on common sense because they were ignored in the annotation process of the original dataset.

To analyze the error cases quantitatively, we visualized the results of MiniCPM-7B, the best open-sourced model, across two discriminative settings, as illustrated in Figure~\ref{confusion}. 
For Y/N questions, the model tends to favor positive responses (i.e., Yes).
Specifically, among all misclassifications, the instances where a \texttt{No} label is incorrectly classified as \texttt{Yes} are twice as times as instances where a \texttt{Yes} label is incorrectly classified as \texttt{No}.
Besides, the model tends to answer \texttt{D} in MCQ settings.
We suspect the tendency is likely due to the imbalance in the distribution of the training data.

\section{ Analysis of Relation Hallucination}
\label{sec:analysis}

To quantitatively compare the decision probability distribution when hallucinations occur, we calculate the average probability of an equal number of relation hallucination and non-hallucination examples, as shown in Table~\ref{average probs}.
\begin{table}[htbp]
\centering
\small
\setlength{\tabcolsep}{10pt}
\begin{tabular}{lcccc}
\toprule
\textbf{Dataset} & \textbf{Reefknot} & \textbf{MMRel} & \textbf{Rbench} \\
\midrule
LLaVA & 0.67 & 0.76 & 0.80 \\
MiniGPT4-v2 & 0.76 & 0.75 & 0.62 \\
\bottomrule
\end{tabular}
\caption{The \textbf{average probability of all hallucination cases}. We evaluate LLaVA and MiniGPT4-v2 on the Reefknot dataset, along with two other representative relation hallucination benchmarks.}

\label{average probs}

\end{table}

The table shows that when hallucinations occur, the confidence level in the answers is quite low. Specifically, experiments conducted on three relational-level datasets indicate that the overall probability of the answers is only about 70\%.
In contrast, under normal circumstances, large language models can achieve probability values of up to 95\% when providing factual and truthful answers.
Therefore, a straightforward approach to detecting relation hallucinations is to utilize the entropy $E(X)$ of the probability distribution.
\begin{equation}
    E(X) = -\sum_{i=1}^{n} p(x_i) \log{p(x_i)}.
\end{equation}

\begin{figure}[t]
    \centering
    \includegraphics[scale=0.5]{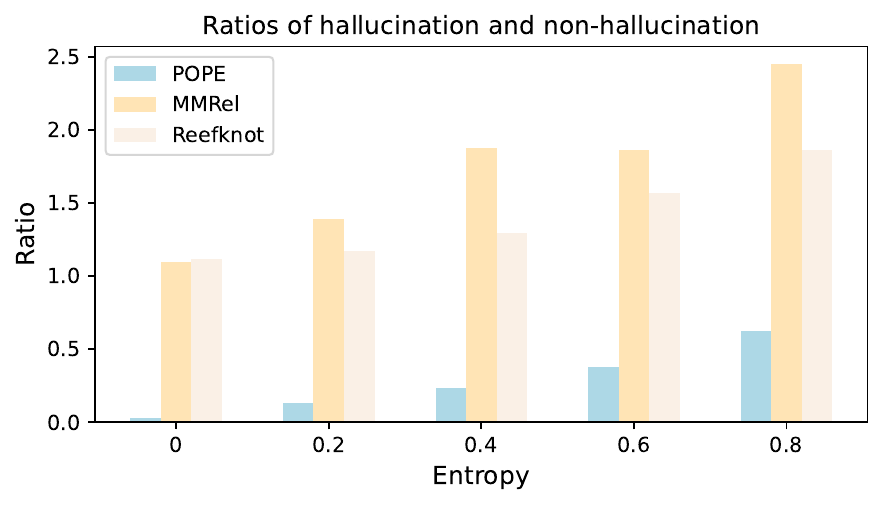}
    \caption{The respective \textbf{ratios between hallucination and non-hallucination} with different entropy values.}%
    \label{entropy ratio}
\end{figure}

\begin{figure}[tbp]
    \centering
    \includegraphics[scale=0.45]{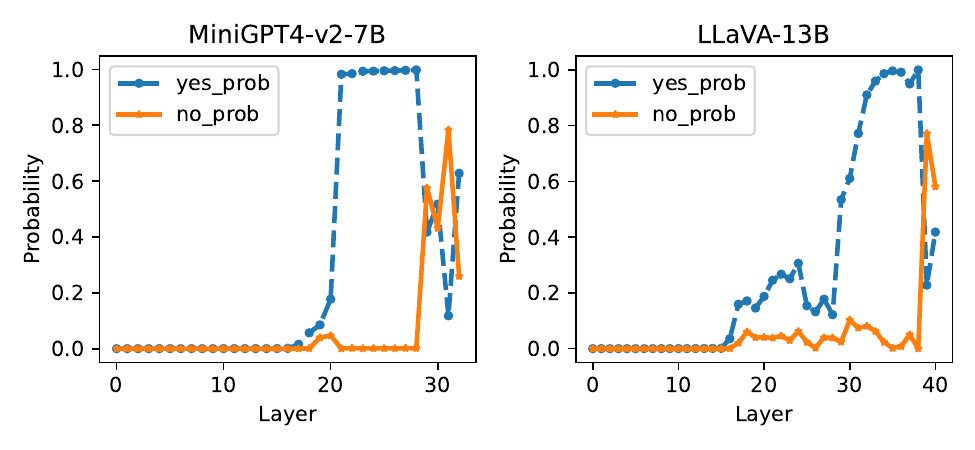}
    \caption{The \textbf{average probability across layers during hallucination occurrences}. Significant probability changes are observed in deeper layers. Results are reported for MiniGPTv2-7B (32 layers) and LLaVA-13B (40 layers) to demonstrate universality.}

    \label{aa}
\end{figure}

Note that because MLLMs have an extensive vocabulary, it becomes challenging to discern meaningful patterns in entropy variation when predicting the next word across the entire distribution.
Consequently, our analysis is restricted to observing the variation patterns of vocabulary within the range of potential answers $x_{i}$.
We present the ratio of hallucination cases to non-hallucination cases among both object and relation hallucination benchmarks in Figure~\ref{entropy ratio}.
When $E(X)>$  0.6, \textit{relation hallucinations} occur to a significant degree, indicating the effectiveness of our method to detect \textit{relation hallucination} via entropy.
To investigate the mechanism behind hallucination generation, we conducted an in-depth analysis of the fluctuations in probability values across model layers.
MLLMs consist of a vision encoder, a projection layer, and a strong LLM decoder, which is stacked by a set of $\mathcal{N}$ transformer layers, and an MLP layer 
 $\phi(\cdot)$ for predicting the next-word probability 
 of distribution auto-regressively.
 Given an image represented by vision encoder as $\mathcal{V}_{t}$ = $\{v_1, v_2, \dots, v_{t}\}$ and a text prompt of tokens $\mathcal{P}_{t}$ = $\{p_1, p_2, \dots, p_{t}\}$, they are processed as a sequence $ \mathcal{H}_{0}$ = $\psi(\mathcal{V}_{t},\mathcal{P}_{t})$
 through projection and concatenation function $\psi(\cdot)$.
Thus $\mathcal{H}_0=\{h_1^{(0)}, \dots, h_{t-1}^{(0)}\}$, in which $h_i^{(k)}$ means the hidden states of $i_{th}$ token in $k_{th}$ language decoder layer.
Then $ \mathcal{H}_{0}$ would be processed by each of the transformer layers in the language decoder successively. 
 We denote the output of the $j$-th layer as $\mathcal{H}_j$. 
 In the normal forward process, $\mathcal{H}_j$ will be calculated by $\mathcal{N}$ times, then it will be passed through language model head layer $\phi(\cdot)$ to predict the probability of the next token $r_{t}$ over the vocabulary set $\mathcal{X}$.
 In our set, we manually pass every $\mathcal{H}_{n}$ to explore the mechanism of hallucination with probability distributions and the next token to generate.   
 For every layer, we can obtain the probability distribution  $\mathbb{P}$ and the next word prediction $r_{t}^{(j)}$. 
\begin{equation}
    \mathbb{P}(\mathcal{H}_{j} \mid \mathcal{H}_{j-1}) = \mathrm{softmax}\bigl(\phi(\mathcal{H}_{j-1})\bigr), \quad j\in \mathcal{N}.
\end{equation}
\begin{equation}
    r_{t}^{(j)} = \arg\max \,  \mathbb{P}(\mathcal{H}_{j} \mid \mathcal{H}_{j-1}^{(j)}) , \quad r_{t}^{(j)} \in \mathcal{X}.
\label{eq:token}
\end{equation}
Using Equation~\ref{eq:token}, we visualized these changes during the forward propagation process, as illustrated in Figure~\ref{aa}. 
To avoid variability from individual instances, we reported the average values of all data involving hallucinations occurring in Reefknot.
To ensure a fair comparison, we utilized the 32-layer MiniGPT4-v2-7B model~\cite{minigptv2} and the 40-layer LLaVA-13B model~\cite{liu2023improved}.

It can be observed that in the shallow layers (0th-20th), the probability of possible answers does not increase.
We hypothesize that this is because, in shallow layers, the model is aggregating information to generate an answer. 
Hallucination occurs in the deep layers. 
Since deep layers contain a vast amount of knowledge, we speculate that they may cause relation hallucinations.

\section{Detect-Then-Calibrate Mitigation Method}
The evaluation results indicate that MLLMs commonly suffer from severe \textit{relation hallucinations}. 
As analyzed in Section~\ref{sec:analysis}, we found that these \textit{relation hallucinations} primarily stem from the model's lack of confidence.
The lack of confidence in the model's responses results in a relatively high entropy value.
Therefore, we can detect the occurrence of  \textit{relation hallucinations} by monitoring changes in entropy.
Initially, we set a specific entropy threshold $\gamma$ to detect potential hallucinations in the model's output. 
If the entropy of the model's response exceeds $\gamma$, which suggests a significant lack of confidence, we infer a high probability that the model has generated hallucinations.
In cases where the model is identified to hallucinate, we will utilize the hidden states from intermediate layers to calibrate the final outputs layers, which is inspired by \citet{dola}. 
Note, unlike traditional contrastive decoding strategies \cite{cd,vcd,opera}, we do not calibrate all cases. 
Instead, we focus on mitigating potential hallucination cases to avoid altering non-hallucinatory cases into hallucination ones. 
Formula~\ref{e:main_one} show our calibration process.

\begin{equation}
\small
    {r}=\left\{\begin{array}{ll}
\arg\max log \frac{(1+\alpha) \cdot softmax(\phi ( h_{t}^{n}))}{\alpha\cdot softmax(\phi ( h_{t}^{n-\lambda}))} , & \text { if }  E_{t} > \gamma. \\ 
\arg\max \bigl(softmax(\phi (\ h_{t}^{n} ))\bigr), & \text { otherwise. }
\end{array}\right.
\label{e:main_one}
\end{equation}

Note that $\lambda$ is the hyperparameter to control the degree of the intermediate layer; \( \mathbf{r} \) is the token generated after calibration; $\alpha$ represents the degree of calibration operation.
The algorithmic flow can be seen from the pseudo-code below:
 \begin{algorithm}[h]
\small
\caption{Detect-Then-Calibrate Algorithm}
\label{alg:training}
\begin{algorithmic}[1]
\REQUIRE Image $v_t$; MLLM $\mathcal{M}(\cdot)$; Prompt $p_{t}$; Uncertainty Entropy threshold $\gamma$
\STATE   ~~~~~~$\mathcal{H}$, $r_0$, $\mathbb{P}$,  =  $\mathcal{M}(x_t,p_t)$  
\STATE The entropy of generate token \\$E(r_0) = -\sum_{i=1}^{n} \rho_i \log{\rho_i}$ $\rho \in \mathbb{P}$
\IF{$E(r_0)$ $\ge$ $\gamma$ }
            \STATE Hallucination occurs! Calibrate $\mathcal{H}$ by first $\lambda$ layer  
            \STATE $h_{\delta}$ = $\log \frac{(1+\alpha)\cdot softmax(\phi (h_{t}^{n}))}{\alpha\cdot softmax(\phi (h_{t}^{n-\lambda}))} $
            \STATE Calibrated response $\bar{r}$ = $\arg\max h_{\delta}$
            \ELSE 
            
            \STATE Normal response $r$ = $\arg\max \bigl(softmax(\phi (\ h_{t}^{n} ))\bigr) $
        \ENDIF
\end{algorithmic}
\end{algorithm}

As illustrated in Table~\ref{alleviate_halluciantion}, we conducted experiments using LLaVA-13B. 
To demonstrate the robustness of our results, we employed two additional relation hallucination datasets, MMRel~\cite{MMREL} and R-bench~\cite{RBench}.
During experiments, we set $\lambda$ = 2 ,$\alpha$ = 0.1, $\gamma=0.9$.
For a fair comparison, in addition to reporting the baseline model, we also report some training-free methods such as VCD~\cite{vcd}, DoLa~\cite{dola}, and OPERA~\cite{opera}. 
Our approach achieved improvements across all three relation hallucination datasets.
Specifically, on the MMRel dataset, our model achieved a 19.7\% improvement compared to the baseline setting.
\begin{table}[h]
\centering
\small
\setlength{\tabcolsep}{2.5pt}
\begin{tabular}{lcccc}
\toprule
\textbf{Methods} & \textbf{Reefknot} & \textbf{MMRel} & \textbf{R-bench} \\
\midrule
Baseline & 37.06 & 40.43 & 29.52 \\
+ VCD~\cite{vcd} & 38.32 & 41.96 & 22.05\\
+ DoLa~\cite{dola} & 36.96 & 39.68 & 23.52\\
+ OPERA~\cite{opera} & 35.73 & 39.22 & 26.73\\
\midrule
\cellcolor{gray!10}+ Detect-then-Calibrate (Ours) & \cellcolor{gray!10}\textbf{34.50} & \cellcolor{gray!10}\textbf{21.73} & \cellcolor{gray!10}\textbf{22.02}\\

\midrule
\end{tabular}
\caption{The \textbf{hallucination rates of LLaVA-13B and its variants}(lower is better) with hallucination mitigation methods across Reefknot and two other relation hallucination benchmarks (\textit{i.e.}, MMRel \& R-bench).}
\label{alleviate_halluciantion}
\end{table}
\section{Case study}
\begin{figure}[h]
    \centering
    \includegraphics[scale=0.25]{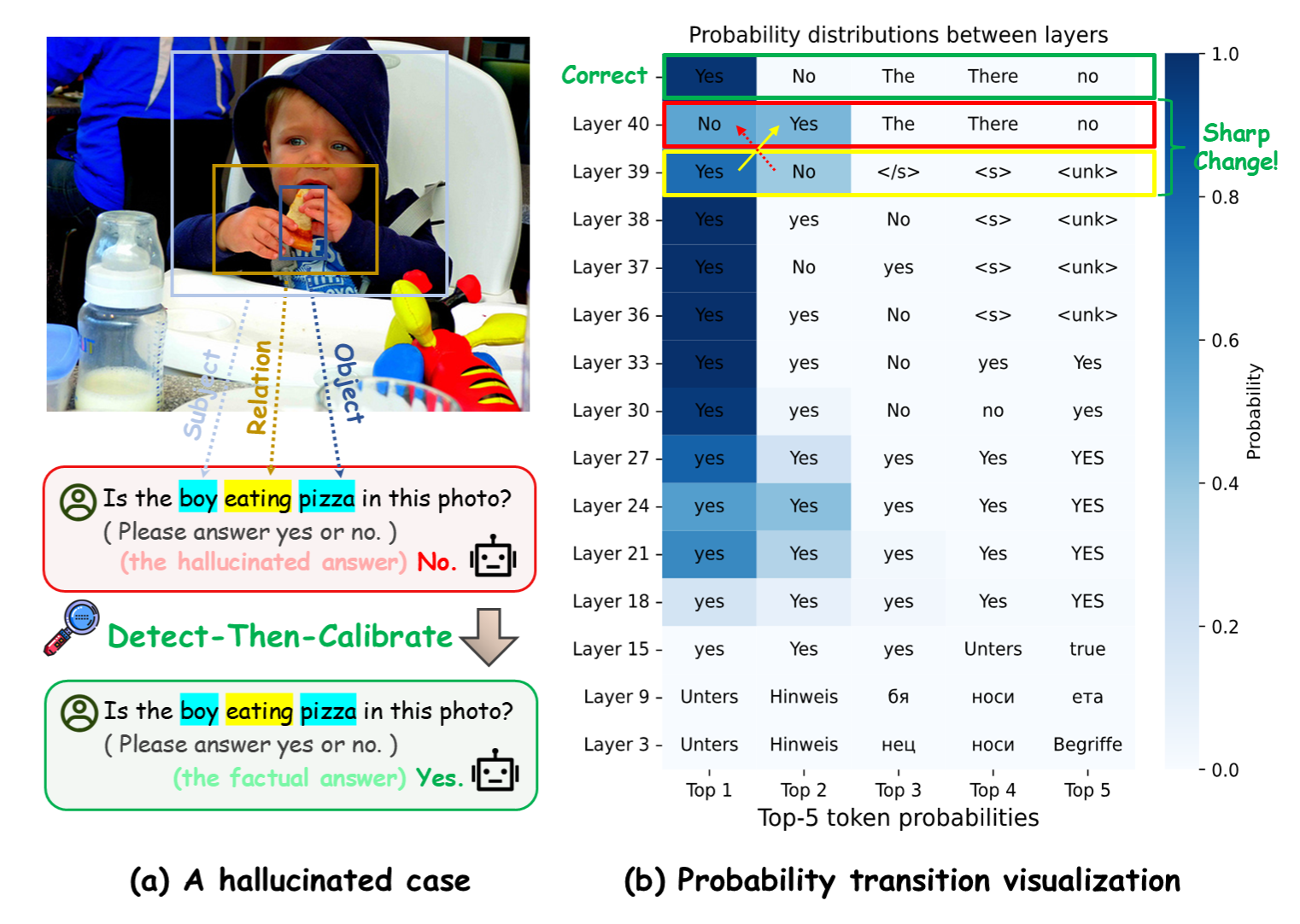}
    \caption{Part (a) illustrates a genuine case of relation hallucination derived from the Reefknot benchmark. The visualization of probability transitions across the layers of the language model is presented in Part (b). Subsequently, our refined results are distinctly highlighted in green, juxtaposed against significant probability variations, which are clearly demarcated within a red-bordered box.}
    \label{case}
\end{figure}

To intuitively demonstrate the generation process of hallucination, we visualized a real case from Reefknot in Figure~\ref{case} \cite{Doe2023}. 
In the image, when we ask ``\texttt{Is the boy eating pizza in the photo?}", MLLMs are unable to provide precise answers.
To investigate this, we analyze the probabilities associated with the top five predicted tokens at each layer of the language model.
As illustrated in Figure~\ref{case}, the model typically converges on the correct answer as it progresses from shallow to deep layers.
Notably, hallucinations didn't occur until the final layer.
Additionally, we can observe that as the number of layers increases, the model increasingly focuses on narrowing down the range of answer choices, with the probability for both `\texttt{yes}' and `\texttt{no}' rising in deeper layers.

However, as depicted in the changes of colors, when it reaches the final decoder block (layer 38-40th), the model's choices suddenly become uncertain and the entropy rises, accompanied by the emergence of hallucinations.
So it is instinctive to utilize the entropy to detect hallucination and use logits of the intermediate layers, which are less likely to hallucinate, to calibrate the final logits to correct the answer.
Following our calibration utilizing Equation~\ref{e:main_one}, we not only dispel potential relation-level hallucination but also significantly bolster the confidence of model's responses.

\section{Related Work}
\paragraph{Hallucination Benchmarks in MLLMs} 
Large Language Models (LLMs) are powerful tools, and exploring their interpretability~\cite{steer,MMneuron}, operational mechanisms, and methods to ensure trustworthy responses\cite{hu2023} has become a crucial area of research.
In the realm of MLLMs, hallucinations are typically categorized into three distinct types: object, attribute, and relation, as outlined in prior studies~\cite{MLLMsurvey,liu2024survey,guan2024hallusionbench}. 
At the object and attribute levels, a considerable number of representative benchmarks such as POPE \cite{POPE} and HaELM \cite{HaELM} have been introduced by researchers.
Many benchmarks are accompanied by a diverse array of evaluation criteria, with generative criteria including CHAIR \cite{CHAIR}, THRONE \cite{THRONE} and various discriminative criteria being particularly notable. 
The existing relation hallucinations benchmarks \cite{RBench, MMREL} focus only on discriminative criteria. The gap of current benchmarks is to identify relationship pairs that accurately reflect the dynamic interactions. 


\paragraph{Confidence Calibration}
Confidence estimation~\cite{zou,chen2024m3cotnovelbenchmarkmultidomain} and calibration are essential for enhancing the reliability of LLMs such as GPT-3~\cite{brown2020language}.
To assess the confidence associated with outputs from LLMs, \citet{SU} have developed a method called semantic entropy that utilizes linguistic invariances reflecting shared meanings. However, this method relies on accessing token-level probabilities, which are often unavailable through current black-box APIs. 
 \citet{know} have designed prompts that encourage the models to self-assess their responses and to explicitly calculate the probability that an answer is true; while \citet{lin} have prompted LLMs to provide both an answer and an accompanying confidence level.

\section{Conclusion }
In conclusion, we propose a comprehensive benchmark called Reefknot to evaluate and mitigate relation hallucinations in MLLMs.
We construct the dataset with over 20k data through a scene graph based construction pipeline, covering two discriminative tasks (Y/N and MCQ) and one generative task (VQA).
Our in-depth evaluation highlights a substantial performance gap on relation hallucination in existing MLLMs, emphasizing the need for more sophisticated reasoning capabilities.
Subsequently, we discover that relation hallucinations tend to occur when MLLMs respond with low confidence.
Therefore, we propose a Detect-then-Calibrate method to mitigate the relation hallucination via entropy threshold, with an average reduction of 9.75\% in the hallucination rate across Reefknot and two other representative relation hallucination datasets.
In general, we anticipate that the Reefknot benchmark we propose will serve as a critical foundation for future developments in the field of trustworthy multimodal intelligence, enabling more reliable and robust systems.
\section{Limitation}
Despite promising, our proposed approach focuses on mitigating basic discriminative hallucinations, but relation hallucinations in open domains are still challenging to quantitatively assess and mitigate. 
In future research, we will delve deeper into the underlying causes of hallucinations in open domains and investigate both the mechanisms and mitigation strategies.
We anticipate that Reefknot will further improve the reliability and practical utility of MLLMs.

\section*{Acknowledgement}
 
This work was supported by Open Project Program of Guangxi Key Laboratory of Digital Infrastructure (Grant Number: GXDIOP2024015); Guangdong Provincial Department of Education Project (Grant No.2024KQNCX028); Scientific Research Projects for the Higher-educational Institutions (Grant No.2024312096), Education Bureau of Guangzhou Municipality; Guangzhou-HKUST(GZ) Joint Funding Program (Grant No.2025A03J3957), Education Bureau of Guangzhou Municipality.

\bibliography{custom}

\appendix

\section{Appendix}
\subsection{Detailed Case Study} 
\begin{figure}[h]
    \centering
    \includegraphics[scale=0.5]{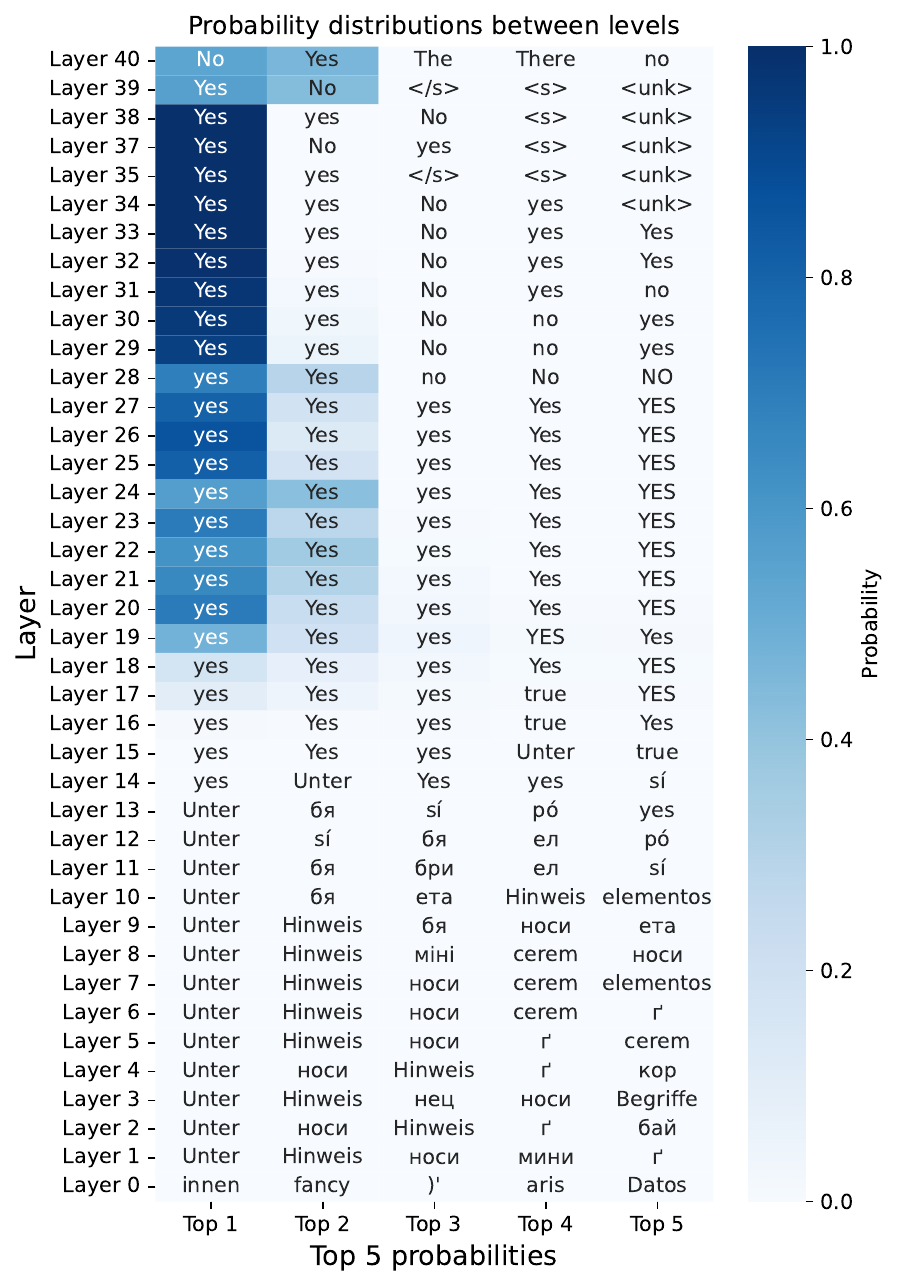}
    \caption{The \textbf{hallucination case} visualization of Top5-token probability transitions across the layers of MLLM.}
    \label{fig:raw_probability_case}
\end{figure}

Due to space limitations, we have provided the complete set of changes in the Top-5 tokens across 40 layers within the Section \textbf{Case Study} in Figure~\ref{fig:raw_probability_case}.

\subsection{Manual Checking Rules}

In session of Triplet Filtering, the following types of relationships need to be avoided:
\begin{enumerate}
    \item \textbf{Redundant and meaningless information}:
    Redundant and meaningless information refers to the relation are overly obvious or self-evident, often involving relationships or facts that are universally understood and require no further explanation. These types of information typically describe natural relation that does not need to be specially emphasized. For example, stating that ‘cows have eyes’ or ‘people wear shirts’ emphasises basic, well-known facts that don't effectively reflect the model's ability to recognise relationships, which is meaningless for our benchmarks.
    \item \textbf{Data Bias}
    In the process of filtering relationships, the number of different relationships should be approximately the same. If the number of cases corresponding to each relationship in the dataset is not balanced, i.e., there are far more cases for some relationships than for others, this will result in a correspondingly higher overall score when the model performs well on a particular relationship. Such a scenario would mask the model's deficiencies in a small number of relationships, leading to distorted assessment results. Therefore, in order to ensure the fairness of the assessment and the reliability of the results, it is important to ensure that the number of cases for each relationship in the dataset is relatively balanced. Otherwise, the strengths and weaknesses of the model will be difficult to accurately reflect through the overall score.
\end{enumerate}

In session of Multi-Turn Manual Checking, the annotator should follow these rules:

\begin{enumerate}
    \item \textbf{Clarity and Precision}
    Ensure that the questions are directly related to verifiable content within the image, steering clear of ambiguity and multiple interpretations. Each question should unambiguously point towards a specific, correct answer.
    
    The following situations need to be avoided:
    \begin{enumerate}
        \item \textbf{Multiple Subjects}: Avoid situations where the presence of multiple subjects in the image leads to unclear references in the semantic triad, making it difficult to discern the subject of the question.
        \item \textbf{Similar Relational Words}: In multiple-choice questions, ensure that the options provided do not include relational words with overlapping or similar meanings.
        \item \textbf{Grammatical Accuracy}: Ensure all samples are free from grammatical errors to maintain the professionalism and clarity of the questions.
    \end{enumerate}
    \item \textbf{Relevance Check}
    Verify that each question is directly related to and necessitates reference to the content of the image. Questions should focus on elements that are prominent and significant within the visual data.
    \begin{enumerate}
        \item \textbf{Minor Details}: Refrain from formulating questions about elements that are minor or indistinct, as they may not significantly influence the understanding of the image.
        \item \textbf{Independence from Image}: Avoid questions that can be answered without referring to the image information, ensuring that the visual content is integral to solving the query.
    \end{enumerate}
\end{enumerate}

Each case is assigned to three annotators for inspection, after which a voting mechanism is used to decide whether or not to keep it in our dataset.
\subsection{Complete guideline for Manual Check}

\textit{Next I will provide you with three types of questions and corresponding images. The types of questions are Multiple Choice Question (MCQ), Yes/No(Y/N), and Visual Question Answer (VQA). MCQ asks about the direct relationship between two objects that already exist in the picture, and the question presupposes four options, with only one option being the final correct answer; Y/N questions presuppose a relationship between the two objects in the image, with the objects already existing in the image; and the VQA is the same as the multiple-choice question setup, but with no options provided. Your task is to make judgments about the reasonableness of the question setup and the relevance of the question to the picture.}

\textit{The degree of reasonableness of the question refers to the fact that the objects involved in the question set must be present in the picture and that the collocation between the two objects has a certain degree of immobility; typical fixed-type collocations are triplets like, <ear on face>, <hair on head>, and when you come across these kinds of relational words in the question set, you should filter them out straight away.}

\textit{The relevance of the picture means that the objects appearing in the question should occupy the main part of the picture, if the picture appears in the minutiae of the question (e.g., when it comes to some very subtle information about the background), you should exclude the picture and the question.}

\textit{You can modify, discard and keep the pairs you see, with the following precautions:}

\begin{enumerate}
    \item[\ding{182}]  \textit{Please exclude unusual and meaningless semantic triplets like <man, wearing, shirt>.}
    \item[\ding{183}]  \textit{Please carefully check the question setting and rewrite some question setting to keep diversity.}
    \item[\ding{184}] \textit{Please make a final judgment on each pair, you can choose to keep, change and keep, or delete.}
\end{enumerate}

\subsection{More Error Analysis}
\begin{figure}[htbp]
    \centering
    \includegraphics[scale=0.5]{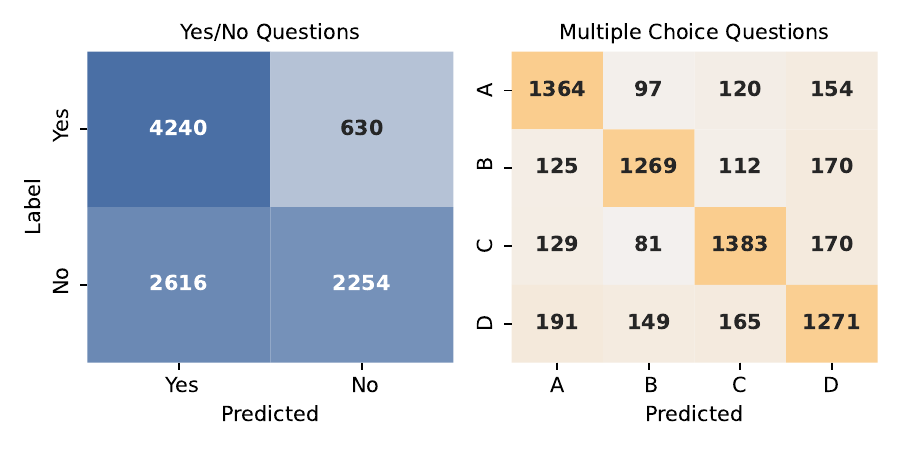}
    \caption{Confusion matrixes of Yi-vl-34b-chat on Reefknot benchmark (Left: Y/N setting; Right: MCQ setting). }%
    \label{fig:34b}
\end{figure}

\begin{figure}[htbp]
    \centering
    \includegraphics[scale=0.5]{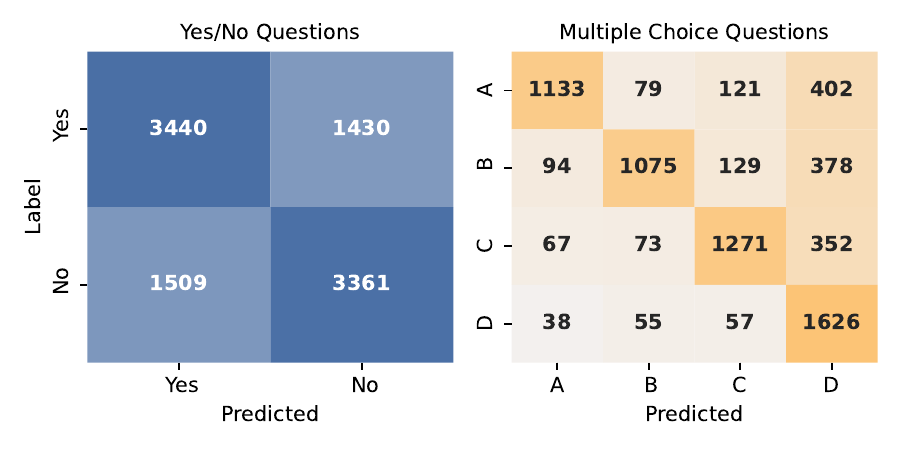}
    \caption{Confusion matrixes of GLM4v-9b-chat on Reefknot benchmark (Left: Y/N setting; Right: MCQ setting). }%
    \label{fig:glm}
\end{figure}

\begin{figure}[htbp]
    \centering
    \includegraphics[scale=0.5]{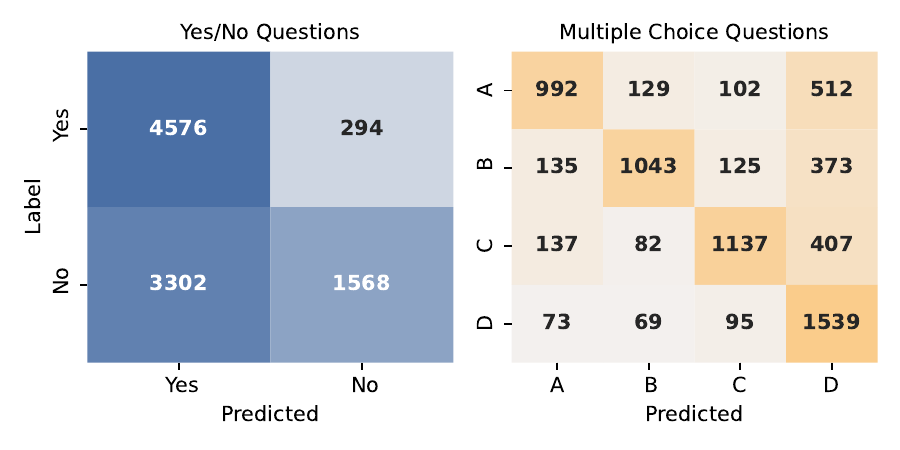}
    \caption{Confusion matrixes of Qwen-vl-chat on Reefknot benchmark (Left: Y/N setting; Right: MCQ setting). }%
    \label{fig:CMqwen-vl-chat}
\end{figure}

\begin{figure}[htbp]
    \centering
    \includegraphics[scale=0.5]{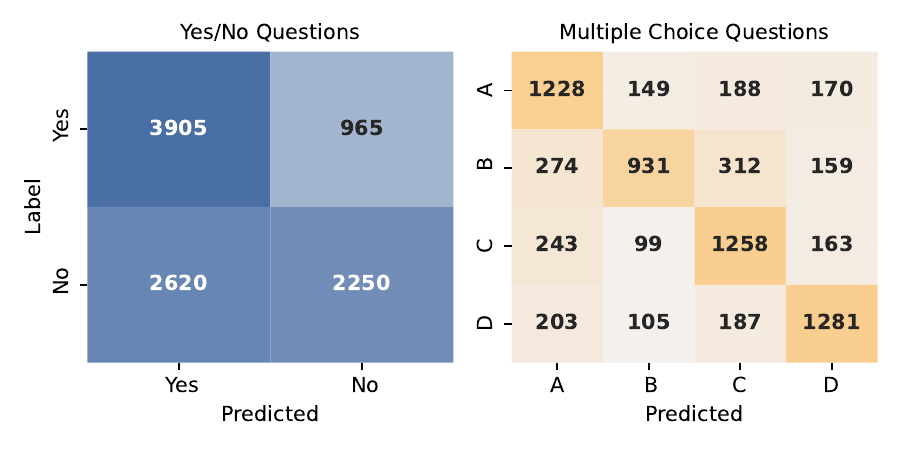}
    \caption{Confusion matrixes of Phi3-vision-128k on Reefknot benchmark (Left: Y/N setting; Right: MCQ setting). }%
    \label{fig:Phi3}
\end{figure}

We listed more error analyses for Y/N and MCQ for MLLMs 
In Figure~\ref{fig:34b}, it shows Yi-VL-34b-chat.
Figure~\ref{fig:glm} shows GLM4v-9b-chat.
Figure~\ref{fig:CMqwen-vl-chat} shows Qwen-vl-chat.
Figure~\ref{fig:Phi3} shows Phi-vision-128k.
It can be observed that, with the exception of Phi-vision-128k, all models demonstrate identical distribution and preference trends.
\newpage
\subsection{Prompt to Categorize}
\label{section prompt}
In Figure \ref{fig:prompt}, we show our prompt used to categorize the relation word used in our benchmark during phase(d) in the pipeline of Reefknot.
\begin{figure}[htbp]
    \centering
\begin{tcolorbox}[colback=white, colframe=black]
\textbf{\textit{Prompt}}: \\
  -------------------------------------------------- \\
You are a relation term classification assistant. 
Please help me determine whether the following relational terms/phrases belong to perceptive relationships or cognitive relationships. \\
Perceptive relationships are defined as those involving locational or state-based prepositions, such as ``on" and ``behind".\\
Cognitive relationships are defined as some words that indicate an action or behavior, such as holding, watching, etc. Here I will give you some demonstrations for reference: 
\textcolor[rgb]{0,0.7,0}{ \{In-context examples\} } \\
Please answer with a single word in \textless \textbf{Perception, Cognition} \textgreater. \\
 Input: \textcolor[rgb]{0.8,0,0}{\{Input\}}\\
\end{tcolorbox}
\caption{Prompt template for relation categorization. } 
\label{fig:prompt}
\end{figure}
\subsection{Potential Concerns}

\textbf{\ding{182} Why detect-than-calibrate method is specific to relation hallucination?}

Our approach is specific to relation hallucinations, which uniquely correlate with high entropy, as shown in Table \ref{average probs} and Figure \ref{aa}. In contrast, other types of hallucinations do not display this trait. The Table \ref{bb} below shows results from the object-level hallucination benchmark POPE \cite{POPE}, where the average probability is over 90\%, resulting in an average entropy of about 0.4, much lower than the 
$\lambda=0.9$ set in the paper.
\begin{table}[htbp]
    \centering
    \begin{tabular}{lcc}
        \toprule
        Model/Dataset & POPE & Entropy \\
        \midrule
        LLaVA & 0.932 & \textbf{0.358} \\
        MiniGPT4-v2 & 0.904 & \textbf{0.456} \\
        \bottomrule
    \end{tabular}
    \caption{The average probability and entropy of object-level hallucination instances.}
    \label{bb}
\end{table}

\textbf{\ding{183} What is the difference between detect-than-calibrate and contrastive decoding based methods like (\textit{i.e.}, vcd , dola)?}

Compared to the \textbf{intuitive} contrastive decoding methods, our approach relies on analyzing relational hallucinations and the variation in hidden layers confidence when relation hallucinations occur. In contrast, the intuitive contrastive decoding method does not conduct a detailed analysis of the issues in the individual cases but instead applies a contrastive-decoding approach to all examples, regardless of whether they are hallucination cases. This may mitigate hallucination effects, while \textbf{leading to non-hallucination examples becoming hallucination ones}, which does not fully address the hallucination. In contrast, our method is based on a cause analysis of relation hallucinations from the perspective of model confidence. We first identified a correlation between relation-level hallucinations and high entropy values. Then, we filter hallucination cases based on entropy values before calibration operation, thereby avoiding the potential negative impacts of contrastive decoding methods on normal instances.

In a word, our approach differs from contrastive decoding methods in that we selectively correct potential hallucination examples based on the characteristics of relation hallucinations. The application of the sole contrastive methods tend to convert many non-hallucination cases into hallucination cases, resulting in no significant effects, as demonstrated in Table \ref{alleviate_halluciantion}.

\textbf{\ding{184}  Why no more experiments on other benchmarks \textit{(i.e., POPE , CHAIR)}?}

POPE~\cite{POPE} and CHAIR~\cite{CHAIR} are indeed well-known benchmarks for hallucination evaluation. However, POPE is specifically focused on \textbf{object-level hallucinations}, while CHAIR is a \textbf{generative metric} for evaluating \textbf{object-level hallucinations} (caption-based).

Our Detect-Than-Calibrate focuses on exploring the mechanisms and mitigation of \textbf{relation-level} hallucinations. In Figure~\ref{entropy ratio}, we demonstrate the hallucination ratio of POPE under different entropy distributions. As shown in Figure~\ref{entropy ratio}, POPE does not show an increase in hallucinations with higher entropy, making it unsuitable for evaluating our entropy-based Detect-than-Calibrate method. CHAIR, on the other hand, is a generative metric for object-level hallucinations, whereas our method targets \textbf{discriminative} relation-level hallucination mitigation. Therefore, it is not feasible to evaluate our approach using POPE and CHAIR.

To validate the effectiveness of our method, besides experiments on our Reefknot, we also conducted experiments on two benchmarks specifically targeting \textbf{relation-level hallucination}, R-bench~\cite{RBench} and MMrel~\cite{MMREL}, in addition to our own Reefknot. The results in Table~\ref{alleviate_halluciantion} demonstrate the effectiveness of our approach. Additionally, in the paper, we explore the distinction between object-level and relation-level hallucinations in Figures~\ref{fig:domain},~\ref{entropy ratio}, and Table~\ref{bb}.

\textbf{\ding{185} The Proof Reliability of DebERTa}

following the setting from~\citet{SU,Farquhar2024}, we reformulated VQA task as an NLI task by appending the instruction ``\emph{Please answer in the following format: Subject is <relation> Object}" to the prompt.
This raises two potential concerns: First, there is a concern about DeBERTa's ability to distinguish between MLLM's response and the ground truth; Second, there is a concern about how the introduction of answer templates might influence the results.

To address the concern regarding DeBERTa's capabilities, we selected a number of examples from the study for verification and had human annotators annotate these examples for comparison, as shown in Table \ref{deberta vs manual}. The results showed that over 95\% of the cases yielded identical annotations between DeBERTa and the human annotators, demonstrating that DeBERTa is well-suited for this task.

To address the concern about the introduction of answer templates, we compared responses with and without templates. Responses generated with the template were structured as phrases, while those without the template were more open-ended and variable. We employed GPT-4o-mini to assess whether the open-ended responses were semantically equivalent to the ground truth. Table~\ref{tab:vsgpt} shows that without a constrained template, MLLMs tend to show a higher hallucination rate, which proves the necessity of the template in prompts. 

\textbf{\ding{186} Why do you set $E(X)>$  0.6 as a threshold to calibrate? The ratio seems unchanged between $E(X)$=0.4 and $E(X)$=0.6 in Figure~\ref{entropy ratio}.}

The higher the entropy value, the lower the confidence of the response, and the greater the likelihood of relation hallucinations occurring. The \(E(X)\) values we use serve as a threshold for our calibration operation. Calibration is performed only when the entropy exceeds this threshold. The benefit of this approach is that it helps avoid unnecessary calibration for cases that do not involve hallucinations. We chose \(E = 0.6\) as a threshold to focus on cases where hallucinations are more likely to occur, \textit{\textbf{ensuring that calibration is applied only to cases with a higher degree of certainty of hallucination, and try our best to reduce calibration on non-hallucination cases}}. 

\textbf{\ding{187} What is the meaning of meaningless questions in manual verification when constructing Reefknot? }

\textit{"Meaningless questions"} are defined as questions like ``Are there ears on a milk cow?'' or ``Are there eyes on a man?'' These types of triplets are fixed combinations, and even without visual information, they can easily be inferred through common sense.

\subsection{VQA Criterion}
\label{vqaCriterion}
\begin{table}[htbp]
    \centering
    \small
    \setlength{\tabcolsep}{3pt} 
    
    \begin{tabular}{lcccc}
        \toprule
        \textbf{Model} & \multicolumn{2}{c}{\textbf{Perception↓}} & \multicolumn{2}{c}{\textbf{Cognition↓}} \\
        \cmidrule(lr){2-3} \cmidrule(lr){4-5}
                      & \textbf{DeBERTa} & \textbf{GPT-assisted} & \textbf{DeBERTa} & \textbf{GPT-assisted} \\
        \midrule
        LLaVA         & 0.68             & 0.93                  & 0.67             & 0.94                  \\
        Deepseek      & 0.70             & 0.89                  & 0.69             & 0.81                  \\
        MiniGPT4      & 0.61             & 0.78                  & 0.61             & 0.83                  \\
        MinCPM        & 0.76             & 0.87                  & 0.69             & 0.94                  \\
        \bottomrule
    \end{tabular}
    \caption{Comparison of Hallucination Rates between DeBERTa-Based and GPT-Assisted Methods in VQA settings.}
    \label{tab:vsgpt}
\end{table}

\begin{figure*}[htbp]
    \centering
    \includegraphics[scale=0.6]{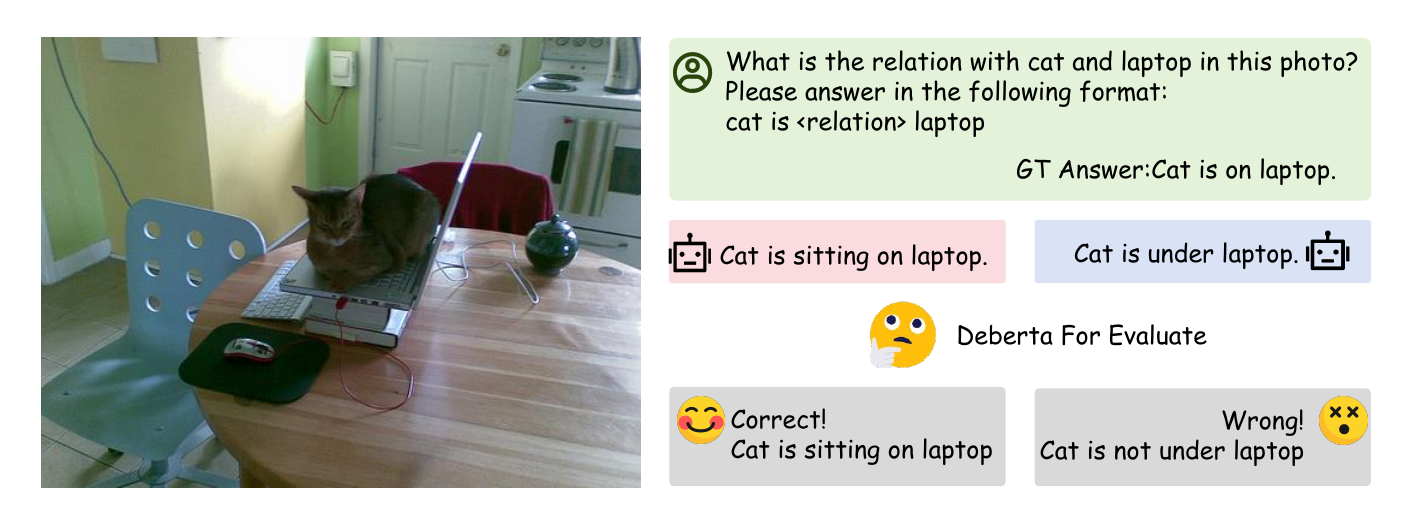}
    \caption{An example of using DeBERTa for evaluation}%
    \label{fig:VQA_case}
\end{figure*}

Here are the evaluation criteria for VQA questions.
We use the DeBERTa\footnote{\url{https://huggingface.co/microsoft/deberta-v2-xlarge-mnli}} model to determine whether the models entail each other.
Only when the label and response contain each other will it be judged as a correct reply. We show our key function code in Figure~\ref{code:are_equivalent}.
To show potential bias associated with the form of constrained prompt and DeBERTa, we compared the evaluation results using a GPT-assisted method. In this setting, no restrictions were imposed on the prompt (e.g., "What is the relationship between A and B?"). The results, as shown in Table \ref{tab:vsgpt}, demonstrate that when the prompt does not constrain the responses, the performance of MLLMs in VQA about relationships is significantly impaired.
To further validate the effectiveness of the DeBERTa model, we compared the judgments made by the DeBERTa with those of human checkers, which demonstrate that DeBERTa can accurately assess the correctness of the model’s responses in Table \ref{deberta vs manual}. 
Besides, we show a VQA case in Figure~\ref{fig:VQA_case}.

\begin{table*}[htbp]
    \centering
    \begin{tabular}{llcc}
        \toprule
        \textbf{Ground truth}       & \textbf{Response}         & \textbf{DeBERTa} & \textbf{Manual check} \\ 
        \midrule
        sunlight is \textbf{shining on} train & sunlight is \textbf{illuminating} train & \blueright       & \blueright            \\ 
        bear is \textbf{reading} book         & bear is \textbf{sitting} on book        & \redwrong       & \redwrong            \\ 
        picture is \textbf{hanging} wall      & picture is \textbf{on} wall             & \blueright       & \blueright            \\ 
        man is \textbf{lying on} couch        & man is \textbf{lying} couch             & \blueright       & \blueright            \\ 
        dog is \textbf{barking} at stranger & dog is \textbf{at} stranger& \redwrong&\redwrong\\
        rain is \textbf{falling on} roof& rain is \textbf{on} roof&\redwrong&\redwrong\\
        car is \textbf{parked in} garage& car is \textbf{in} garage&\blueright&\blueright\\
        fish is \textbf{swimming} pond & fish is \textbf{darting} water&\redwrong&\redwrong\\
        dog is \textbf{barking} yard& dog is \textbf{howling} garden&\blueright&\blueright\\
        fire is \textbf{burning} hearth&fire is\textbf{ crackling} fireplace&\blueright&\blueright\\

        \bottomrule
    \end{tabular}
    \caption{The comparison of results between DeBERTa and Manual Check for real cases from our Reefknot.}
    \label{deberta vs manual}
\end{table*}

\subsection{More Cases}
\begin{figure*}[htbp]
\centering
\includegraphics[width=1\textwidth]{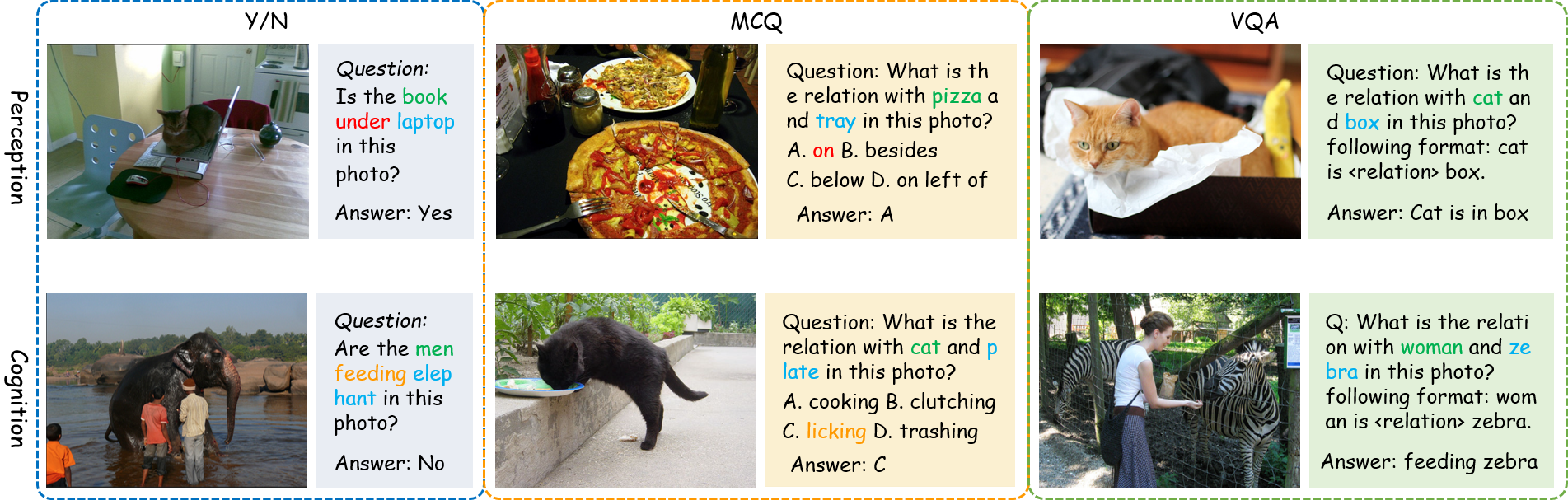}
\caption{Real-world cases from our proposed Reefknof benchmark. We outline the questions for three types of tasks between perception and cognition perspectives as a reference.}
\label{fig:full case}
\end{figure*}

Figure~\ref{fig:full case} presents examples of Yes/No, multiple-choice questions (MCQ) and visual question answering (VQA) tasks under perceptive and cognitive conditions, shown separately for comparison.

\subsection{Visualization of Relation Word}
\begin{figure*}[!t]

    \includegraphics[scale=0.7]{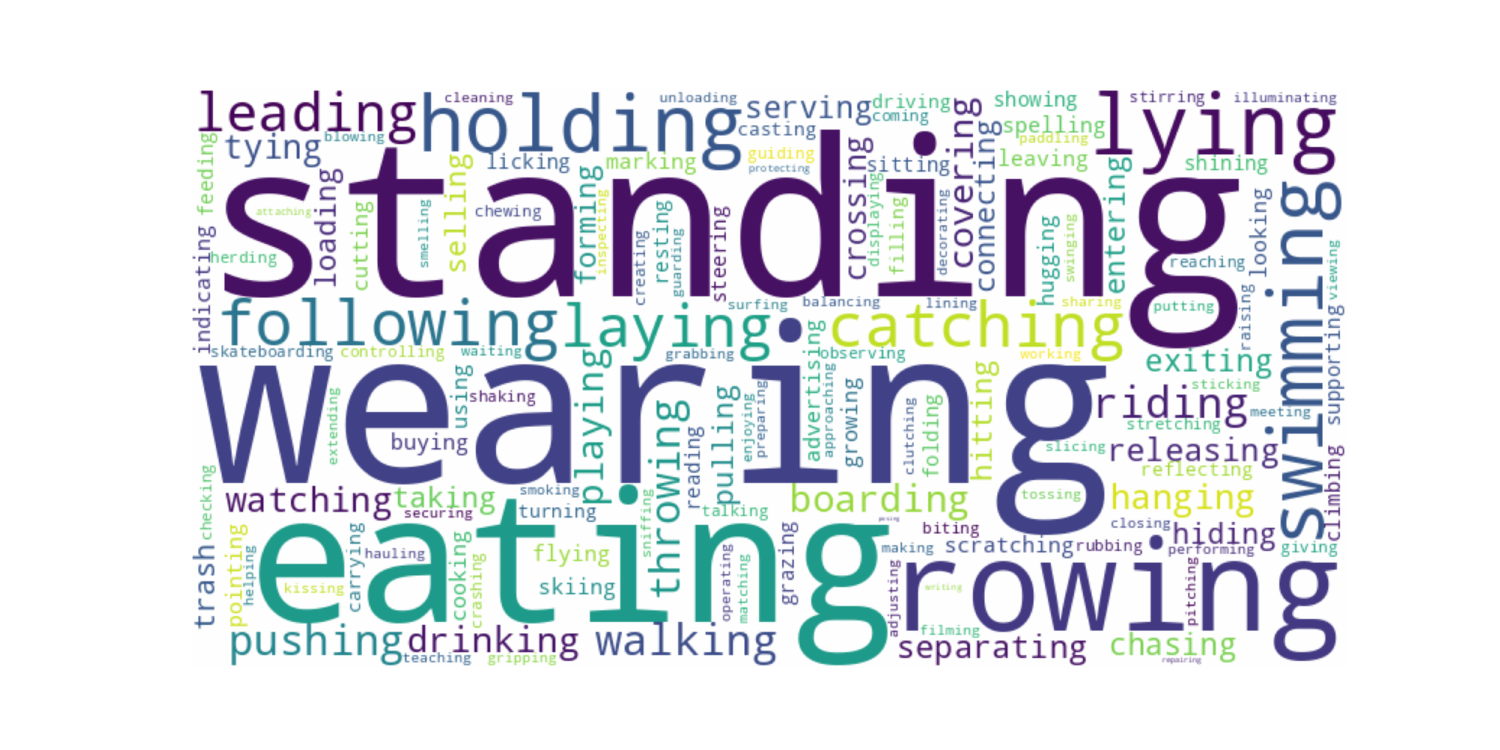}
    \caption{The word cloud of our relations on Reefknot.}
    \label{fig:wordcloud}
\end{figure*}

In Figure~\ref{fig:wordcloud}, we present a word cloud that visualizes the proportion of relational terms within our dataset. It can be observed that, due to the use of semantic triples, our relational terms exhibit greater diversity.
\begin{figure*}[]
\centering
\begin{lstlisting}[language=python, keywordstyle=\color{blue!70}, commentstyle=\color{red!50!green!50!blue!50}, frame=shadowbox, rulesepcolor=\color{red!20!green!20!blue!20}]
 
def are_equivalent(label,response , model, tokenizer, device):
        #label is groundtruth; response is the reply of VLM; 
        def check_implication(label, response):
            inputs = tokenizer(label,response, return_tensors="pt").to(device)
            outputs = model(**inputs)
            logits = outputs.logits
            largest_index = torch.argmax(F.softmax(logits, dim=1))
            return largest_index.cpu().item()
        implication_1 = check_implication(label, response)
        implication_2 = check_implication(response, label)

        assert (implication_1 in [0, 1, 2]) and (implication_2 in [0, 1, 2])
        implications = [implication_1, implication_2]
        semantically_equivalent = (implications[0] == 2) and (implications[1] == 2)
        # only when both are 2, they are semantically equivalent
        return "yes" if semantically_equivalent else "no"
\end{lstlisting}
\caption{Function to check the semantic equivalence of response and our label.}
\label{code:are_equivalent}
\end{figure*}
\clearpage

\end{document}